\documentclass[10pt,twocolumn,letterpaper]{article}

\usepackage{cvpr}
\usepackage{times}
\usepackage{epsfig}
\usepackage{graphicx}
\usepackage{amsmath}
\usepackage{color}
\usepackage[table]{xcolor}
\usepackage{hhline}
\usepackage{cite}
\usepackage{stfloats}
\definecolor{Gray}{gray}{0.85}
\usepackage[utf8]{inputenc}

\newcommand{\depth}{
\begin{figure*}[t]
    \centering
    \includegraphics[width=1\linewidth,{trim=0in .5in 0in 0in,clip=true}]{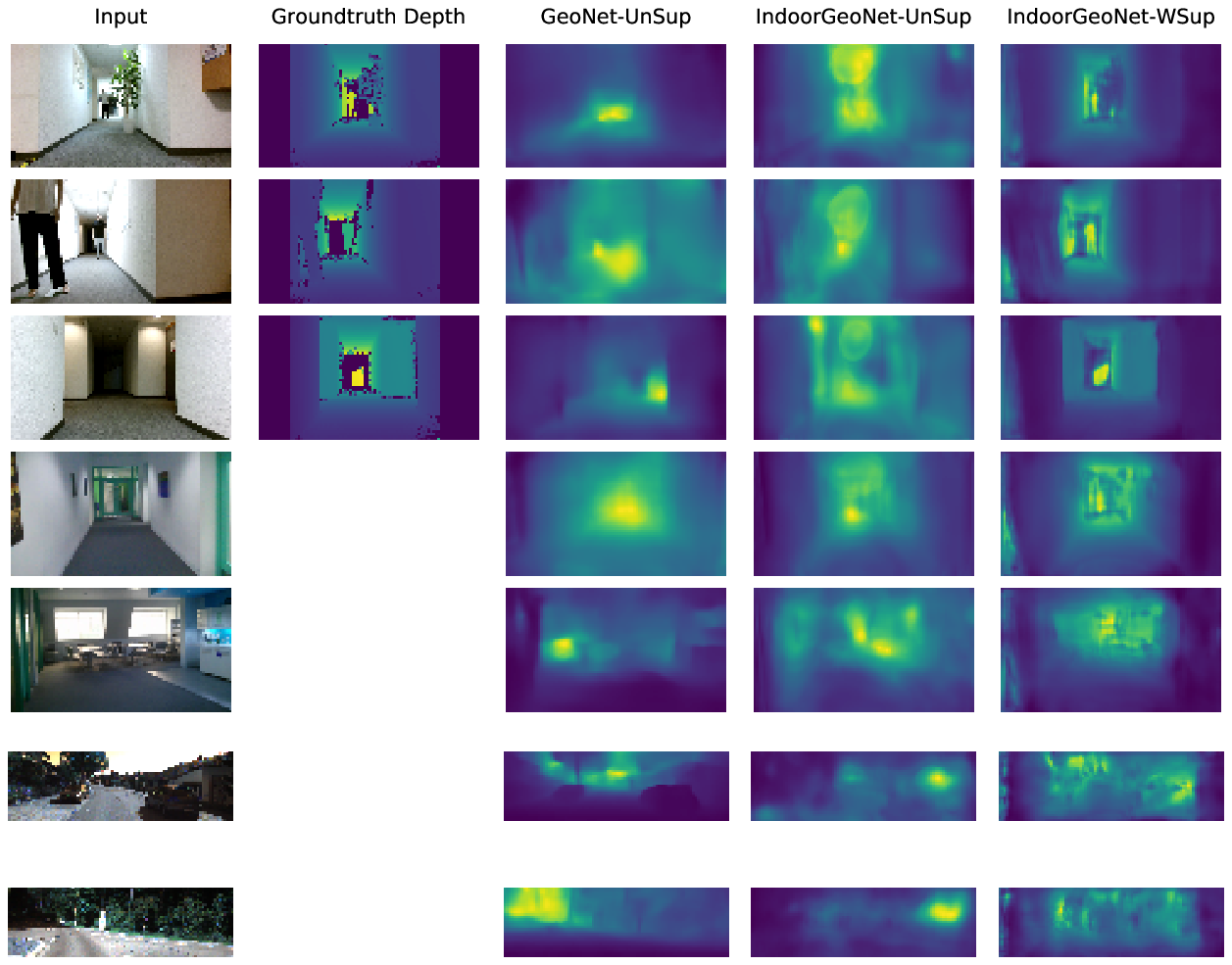}
    \caption{Sample examples of depth image prediction comparing weakly supervised Indoor GoeNet (IndoorGeoNet-WSup) with the original GeoNet trained on KITTI raw and odometry datasets (GeoNet-UnSup) and the GeoNet trained from scratch on the entire RSM Hallway dataset (IndoorGeoNet-UnSup). The rows 1-3 are samples from MobileRGBD, rows 4-5 are from RSM Hallway, and rows 6-7 are from KITTI datasets.}
    \label{fig:depth}
    \vspace{-.3cm}
\end{figure*}
}

\newcommand{\reconst}{
\begin{figure*}[t]
    \centering
    \includegraphics[width=1\linewidth,{trim=0in .6in 0in 0in,clip=true}]{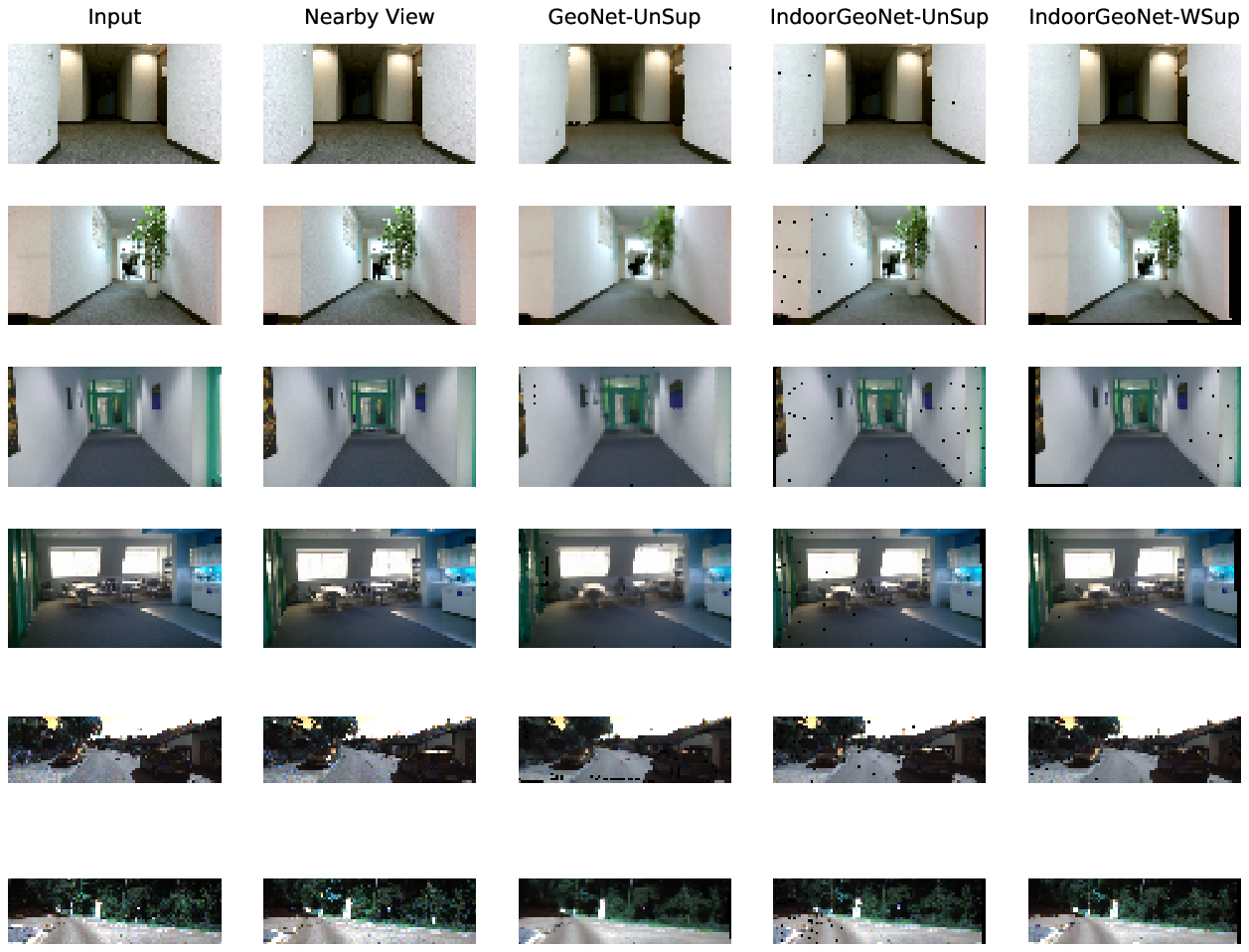}
    \caption{Novel view synthesis samples comparing the reconstruction results of using weakly supervised Indoor GoeNet (IndoorGeoNet-WSup) with GeoNet-UnSup and IndoorGeoNet-UnSup models. The rows 1-2 are samples from MobileRGBD, rows 3-4 are from RSM Hallway, and rows 5-6 are from KITTI datasets.}
    \label{fig:const}
 \vspace{-.2cm}
\end{figure*}

}

\newcommand{\tablecomparedepth}{
\begin{table*}[]
\centering
\caption{Depth and relative camera pose estimation performance comparison between different methods.}
\begin{center}
\label{tbl:depth}
\footnotesize
\begin{tabular}{|*6{p{20mm}|}}
\hline
Method & Training Dataset & \multicolumn{2}{c|}{Depth RMSE} & \multicolumn{2}{c|}{Pose RMSE}\\
\cline{3-6}
 & & KITTI Raw & MobileRGBD & KITTI Odometry & MobileRGBD\\
\hline
GeoNet-UnSup & KITTI & 4.01 & 1.24 & 0.012 & 0.042\\
\hline
IndoorGeoNet-UnSup & RSM Hallway & 13.37 & 1.14 & 0.057 & 0.034\\
\hline
\textbf{IndoorGeoNet-WSup} & MobileRGBD & 12.82 & \textbf{0.72} & 0.051 & \textbf{0.006} \\
\hline
\end{tabular}
    \vspace{-.7cm}
\end{center}
\end{table*}

}

\newcommand{\tablecomparerecons}{
\begin{table*}[b] 
\centering
\vspace{-.1cm}
\caption{Novel view synthesis performance comparison between different methods. Please note that while reconstruction error (RMSE) is desired to be low, the reconstruction similarity measure (SSIM) is desired to be close to 1.}
\begin{center}
\label{tbl:reconst}
\footnotesize
\begin{tabular}{|*8{p{17mm}|}}
\hline
Method&Training Dataset&\multicolumn{3}{c|}{Reconstruction $\mathcal{L}_{P}$ Loss (RMSE)}&\multicolumn{3}{c|}{Reconstruction Similarity (SSIM)}\\
\cline{3-8}
 & & KITTI & RSM Hallway & MobileRGBD & KITTI & RSM hallway & Mobile RGBD\\
\hline
GeoNet-UnSup & KITTI & 32.71 & 28.65 & 32.92 & 0.29 & 0.12 & 0.20\\
\hline
IndoorGeoNet-UnSup & RSM Hallway & 50.20 & 18.62 & 26.94 & 0.21 & 0.40 & 0.43\\
\hline
\textbf{IndoorGeoNet-WSup} & MobileRGBD & 47.42 & 23.74 & \textbf{21.2} & 0.24 & \textbf{0.38} & \textbf{0.49}\\
\hline
\end{tabular}
    \vspace{-.7cm}
\end{center}
\end{table*}
}

\newcommand{\eqnref}[1]{Eq.~(\ref{eqn:#1})}
\newcommand{\figref}[1]{Fig.~\ref{fig:#1}}
\newcommand{\tblref}[1]{Table~\ref{tbl:#1}}

\usepackage{multirow}

\usepackage[pagebackref=true,breaklinks=true,letterpaper=true,colorlinks,bookmarks=false]{hyperref}

 \cvprfinalcopy 

\def\cvprPaperID{3685} 

\ifcvprfinal\fi
\begin{document}

\title{Indoor GeoNet: \\
Weakly Supervised Hybrid Learning for Depth and Pose Estimation}

\author{Amirreza Farnoosh\\
Northeastern University\\
Boston, MA, USA\\
{\tt\small farnoosh.a@husky.neu.edu}
\and
Sarah Ostadabbas\\
Northeastern University\\
Boston, MA, USA\\
{\tt\small ostadabbas@ece.neu.edu}
}

\maketitle

\begin{abstract}
Humans naturally perceive a 3D scene in front of them through accumulation of information obtained from multiple interconnected projections of the scene and by interpreting their correspondence. This phenomenon has inspired artificial intelligence models to extract the depth and view angle of the observed scene by modeling the correspondence between different views of that scene. 
Our paper is built upon previous works in the field of unsupervised depth and relative camera pose estimation from temporal consecutive video frames using deep learning (DL) models. Our approach uses a hybrid learning framework introduced in a recent work called GeoNet, which leverages geometric constraints in the 3D scenes to synthesize a novel view from intermediate DL-based predicted depth and relative pose. However, the state-of-the-art unsupervised depth and pose estimation DL models are exclusively trained/tested on a few available outdoor scene datasets and we have shown they are hardly transferable to new scenes, especially from indoor environments, in which estimation requires higher precision and dealing with probable occlusions. This paper introduces ``Indoor GeoNet'', a weakly supervised  depth and camera pose estimation model targeted for indoor scenes. In Indoor GeoNet, we take advantage of the availability of indoor RGBD datasets collected by human or robot navigators, and added partial (i.e. weak) supervision in depth training into the model. Experimental results showed that our model effectively generalizes to new scenes from different buildings. Indoor GeoNet demonstrated significant depth and pose estimation error reduction when compared to the original GeoNet, while showing 3 times more reconstruction accuracy in synthesizing novel views in indoor environments.    
\end{abstract}

\vspace{-.2cm}

\section{Introduction}
In spite of extensive research in the field of indoor navigation, this problem is still unsolved \cite{huang2009survey}. In order to travel a path in an unknown or even known indoor scene a map along with a positioning system needs to be provided to the navigator (e.g. a person or a robot). The global navigation satellite system (GNSS) data collected over time provides such information in outdoor scenes \cite{ojeda2007personal}. However, GNSS signals are usually not available or very weak in indoor places. To compensate for the lack of GNSS inside buildings, information from other sensing modalities such as artificially installed beacons or wearable inertial measurement units (IMUs) are often used for odometry \cite{farnoosh2018first}. However, data from these sources are usually not reliable over an extended period of time due to the extensive drift caused by accumulation error \cite{woodman2007introduction}. Besides that, these sensors can only provide low-level information about the scene, and are not able to reveal any other information about the overall 3D structure of the indoor places to be used for scene understanding, dynamic interaction, and ultimately a reliable indoor navigation.

In contrast to the sparse distance-based sensing, information such as depth and relative camera pose can be used together to give a very accurate and detailed representation of an indoor scene \cite{izadi2011kinectfusion}. These information could facilitate both navigation and dynamic interaction and also help to reconstruct a unified 3D model of the scene for the purpose of map generation of unknown places \cite{maier2017efficient,dzitsiuk2017noising} or even adding augmented reality features to the scene for a better interaction experience \cite{izadi2011kinectfusion}. It can also be used for virtual tours of an indoor scene while the observer looks into the rendered scenes in different views \cite{eslami2018neural}.

In computer vision field, there has been extensive research for indoor odometry, scene understanding, and specifically camera pose, depth, and flow extraction from a moving camera (e.g. robot or head mounted), most of which are recently powered with the advances in deep learning (DL) used in visual simultaneous localization and mapping (vSLAM) works \cite{cadena2016past}. The use of DL in vSLAM can be separated into two categories of supervised and unsupervised learning, while the former is more studied \cite{cadena2016past}. The availability of the open-source benchmark datasets from outdoor scenes (e.g. KITTI \cite{menze2015object}) or indoor scenes (e.g. NYU Depth \cite{silberman2012indoor}, RSM Hallway Dataset 
\footnote{http://www.bicv.org/datasets/rsm-dataset/}, and MobileRGBD \footnote{http://mobilergbd.inrialpes.fr/}) has been very crucial in the success of DL-based supervised vSLAM models. However, similar to the most cases in supervised learning, the need for large and diverse data/label pairs for training such deep networks is still a limiting factor in this domain. Recently, there have been several works proposed for unsupervised learning of depth and camera pose from video frames that use time order of consequent video frames as their hidden supervision signal, including the GeoNet presented in CVPR2018 \cite{yin2018geonet}. In particular, GeoNet took advantage of a hybrid learning approach by combining an unsupervised deep learning algorithm and a geometry-based reconstruction equation into a same inference framework. This hybrid learning approach allows to integrate the domain knowledge (i.e. 3D scene geometry constraints) into the framework to suppress physically unfeasible solutions. Although, such unsupervised configurations can be trained on any amount of data without labeling cost, they still fall behind supervised methods in terms of the estimation accuracy, and are hardly generalizable to new scenes which are not seen \emph{apriori} by the network. 

Inspired by the GeoNet framework, in this paper we propose our ``Indoor GeoNet'' model, which is a \emph{weakly} supervised hybrid learning approach for camera pose, depth and flow estimation targeted to indoor scenes. Capitalized on the availability of the inexpensive depth sensing (e.g. Microsoft Kinect and Intel RealSense), we introduce the weak supervision by providing a set of groundtruth depth data into the model during the training. This type of supervision is weak due to the fact that the model is only partially supervised on depth data and the camera pose needs to be learned in an unsupervised fashion. We also believe that this kind of weak supervision for indoor scene understanding has recently become viable since the release of several RGBD open-source datasets collected by human or robot navigators such as NYU Depth, and MobileRGBD datasets.









\section{Related Work}

\subsection{Supervised Approach to Deep Learning of Scene}


In the last few years, there has been several studies for supervised learning of the depth and camera pose \cite{cadena2016past}. In one of the early works \cite{eigen2014depth}, Eigen et al. proposed a two-level network architecture, in which a coarse global prediction from the first stage was refined locally by a fine-scale network thereafter.
This network was trained using a scale-invariant error that compares the final lower-resolution output with its corresponding groundtruth map, and could achieve state-of-the-art results on both NYU Depth and KITTI datasets.





In another work by Fischer et.al \cite{fischer2015flownet}, authors proposed an encoder-decoder architecture for flow  prediction in an end-to-end training fashion given datasets consisting of image pairs and their corresponding groundtruth flows. In a follow-up study \cite{ilg2017flownet}, the authors realized that the network performance could  be improved if the training data with different properties are presented to the network. Additionally, they proposed a stacked architecture that takes warping of the second image with intermediate optical flow as input for further refinement, as well as a sub-network for improving prediction on small motions, which could obtain state-of-the-art results on a few benchmark datasets, including Sintel \footnote{http://sintel.is.tue.mpg.de/}, Middlebury \footnote{http://vision.middlebury.edu/stereo/data/}, and KITTI datasets.

Early 2018, Liu et al. addressed the problem of novel view synthesis from a single image using an architecture with two sub-networks \cite{liu2018geometry}. 
One of the sub-networks, adapted from the work of \cite{eigen2015predicting},
is responsible for pixel-wise prediction of depth and normals from a single image, which was pre-trained and fine-tuned in a supervised manner. These predictions together with the extracted region masks and relative poses are then used by the second sub-network to compute multiple homographies to warp input image into a novel view. The entire network is finally trained end-to-end on pairs of images.

In another recent work, Eslami et al. introduced a probabilistic generative network for 3D rendering of a scene given multiple viewpoints \cite{eslami2018neural}. This network takes as input images of a scene taken from different viewpoints and their corresponding camera poses, constructs an internal representation, and uses this representation to predict novel views from new query viewpoints. It is trained on pairs of images and their corresponding viewpoints from millions of synthetic  scenes. Although, they obtained promising results for these synthetic scenes, they faced computational difficulties when implementing their network on real datasets.





\subsection{Unsupervised Approach to Deep Learning of Scene}
The unsupervised learning of the depth and camera pose could alleviate or remove the need for expensive data acquisition and labeling process. The common methodology behind all of the recent unsupervised methods for vSLAM is warping one image in pairs of related images (either stereo pairs or consecutive frames in a video) to the other view by leveraging the geometry constraints of the problem, in an approach very similar to the idea of autoencoders.


In one of the earlier work in this topic, Garg et al. proposed an unsupervised deep convolutional neural network (CNN) for single view depth prediction \cite{garg2016unsupervised}. At training time, they considered pairs of stereo images, and trained the network by warping one view to the other one using the intermediate predicted depth and known inter-view displacement through an image similarity loss. They used Taylor expansion of the geometric warping function to make it differentiable for neural network training.
Later, Godard et al. showed that this photometric loss combined with a consistency loss between the disparities produced relative to both the left and right images of a stereo pair, would lead to improved performance and robustness in depth prediction \cite{godard2017unsupervised}.


In another concurrent work, Jason et al. proposed an unsupervised approach to train a CNN for predicting optical flow between two images \cite{jason2016back}. The network was trained using pairs of temporally consecutive images through a photometric loss between the first image and the inverse warping of the second image as well as a flow smoothness loss term. In a similar approach, Vijayanarasimhan et al. proposed a geometry-aware  network that predicts depth, camera pose, and a set of motion masks corresponding to rigid object motions and segmentation given a sequence of consecutive frames in the input \cite{vijayanarasimhan2017sfm}. This is performed by converting the predictions to optical flow and then warping the frames to each other and considering forward-backward consistency constraints.







Following the same approach, Zhou et al. proposed a network for jointly unsupervised training of a depth CNN and a camera pose estimation network from video sequences in mostly rigid scenes again by leveraging a photometric loss from novel-view synthesis \cite{zhou2017unsupervised}. They additionally trained an explainability prediction network which outputs a per-pixel soft mask, with which the view synthesis objective is weighted, in order to handle visibility, non-rigidity and other non-modeled factors.
Meister et al. proposed using a robust census transform for the photometric loss along with an occlusion-aware loss to mask occluded pixels, whenever there is a large mismatch between estimated forward and backward flows \cite{meister2017unflow}.

In October 2018, Ranjan et al. released their work, which used an adversarial collaboration structure for jointly unsupervised learning of depth, camera motion, optical flow, and motion segmentation from video sequences \cite{ranjan2018adversarial}. They used two adversaries in their framework, one for static scene reconstruction based on estimated depth and camera motion, and one for moving region reconstructor based on estimated flow. This competition is moderated by a pixel-wise probabilistic motion segmentation network that distributes training data to these adversaries. The moderator itself is trained to segment static and moving regions correctly by taking a consensus between flow of the static scene and moving regions from the two adversaries. The authors argue that since these four fundamental vision problems are coupled, learning them together would result in an enhanced performance.

Last but not least, Zhichao et al. proposed GeoNet, a jointly unsupervised deep network for depth, camera pose and flow estimation given a sequence of video frames \cite{yin2018geonet}. They broke down the problem of flow estimation into two parts: rigid flow which handles static background, and non-rigid flow which handles moving objects, and assigned two cascaded sub-networks to perform full flow estimation accordingly. In addition, they proposed an adaptive geometric consistency loss inspired by \cite{godard2017unsupervised} to increase robustness towards outliers and non-Lambertian regions. 

In this paper, we utilized the hybrid learning framework of GeoNet, which is trained and tested on videos from outdoor scenes. Specifically, these videos are recorded from a fixed camera mounted on top of a car. We argue that this approach is not well transferable to indoor scenes for several reasons: Firstly, in contrast to outdoor scenes, the relative displacement with respect to the depth range is high in the indoor scenes. Besides that, the outdoor scenes are much wider, and therefore are less affected by camera movement. Secondly, the relative range of camera pose angles is much less in outdoor scenes, however for indoor scenes, because of the limited space, the changes in camera view can be much sharper. In addition, the head mounted cameras are much more affected by distorted movement. Thirdly, depth precision needed for indoor scenes is much higher because of shorter depth ranges in a more detailed scene with more edges. Our proposed ``Indoor GeoNet''  addressed these issues and provides an efficient hybrid learning framework for accurate pose and depth estimation in indoor scene, based on weak supervision to exploit the advantages of supervised and unsupervised learning in a unified framework.



\section{Building the Indoor GeoNet}
\label{sec:IndoorGeoNet}
Indoor GeoNet shares the same network structure as original GeoNet \cite{yin2018geonet}, in which two sub-networks called DepthNet and PoseNet predict rigid layout of the observed scene including the depth and relative camera pose. The training samples to the network are temporal consecutive frames $I_i(i = 1 \sim N)$ for $N=3$ or $N=5$ with known camera intrinsics. Typically in a sequence of frames, a reference frame $I_r$ is specified as the reference view, and the other frames are target frames $I_t$. 
During training, the DepthNet takes the entire sequence concatenated along batch dimension as input. This allows for single view depth prediction at the test time. 
In contrast, the PoseNet is naturally fed with the entire sequence concatenated along channel dimension, and outputs all of the relative camera poses. This allows the network to learn the connections between different views in a sequence. Fused with the deep structures of DepthNet and PoseNet, rigid scene geometry equations then will be used to warp a target view to the reference view. Unlike the fully unsupervised approach of original GeoNet, Indoor GeoNet takes advantage of the depth supervision to enhance the transferability of the pose and depth learning to indoor scenes.  

\subsection{Geometric-Based Hybrid Learning}

Static scene geometry can be well-defined from patterns of motion of objects, surfaces, and edges in a sequence of ordered images collected from a visual scene. This scene level consistent movement perceived in image plane, known as optical flow, is governed by the relative camera motion between an observer and a scene.  
Therefore, this rigid optical flow can be completely modeled 
by a collection of depth maps $D_i$ for frame $I_i$, and the relative camera motion $\mathbf{T}_{r\rightarrow t}=[R|T]$ from reference frame $I_r$ to target frame $I_t$, where $R_{3\times3}$ and $T_{3\times1}$ represent the relative rotation and displacement matrices, respectively. 
Let $p_r=[X, Y, 1]^T$ denote the homogeneous coordinates of a pixel in the reference view, $D_{p_r}$ be its depth value, $[x, y , D_{p_r}] ^T$ be its corresponding 3D coordinates (referenced on camera pinhole), and $K_{3\times3}$ be the camera intrinsic matrix. Then, $p_r$ in the image plane is:

\begin{equation}
\label{eqn:reference}
p_r = \begin{bmatrix}X\\Y\\1\end{bmatrix} = \frac{1}{D_{p_r}} \mathbf{K} \begin{bmatrix}1 & 0 & 0|0\\0 & 1 & 0|0\\0 & 0 & 1|0\end{bmatrix}
        \begin{bmatrix}x\\y\\D_{p_r}\\1\end{bmatrix}\\  
\end{equation}

Moreover, we can  obtain the projected coordinates of $p_r$ onto the target view $p_t$, as:
\begin{equation}
\label{eqn:projection}
p_t \sim \mathbf{K} \mathbf{T}_{r\rightarrow t}
        \begin{bmatrix}x\\y\\D_{p_r}\\1\end{bmatrix} =  \mathbf{K} [R|T] \begin{bmatrix}x\\y\\D_{p_r}\\1\end{bmatrix}  
\end{equation}

Rewriting the \eqnref{reference} will result in $\begin{bmatrix}x\\y\\D_{p_r}\end{bmatrix}= D_{p_r} \mathbf{K}^{-1}p_r$ and merging that with the \eqnref{projection} will give us the corresponding target pixel coordinates $p_t$ in terms of the reference depth map $D_{p_r}$, reference pixel coordinates $p_r$, and the relative camera motion $[R|T]$, as: 

\begin{equation}
\label{eqn:geometry}
        p_t \sim \mathbf{K} \Big[D_{p_r}  R\mathbf{K}^{-1} p_r + T \Big]
\end{equation}

Using \eqnref{geometry}, we can synthesize a novel nearby view from a reference frame in non-occluded regions having the depth map of pixels in the reference view as well as the relative camera pose between the views. Therefore, the DepthNet and PoseNet can be trained together through novel view synthesis between any pairs of training samples.

\subsection{Weakly Supervised Multi-Objective Training}
Let us denote consecutive frames $\{I_1, \dots, I_r, \dots,  I_N\}$ as a training sequence with the middle frame $I_r$ being the reference view and the rest being 
the target views, $I_t$'s. Then, $\hat{I}_{t\rightarrow r}$ represents the target view
$I_t$ warped to the reference coordinate frame by taking the predicted depth $\hat{D}_r$, the predicted camera transformation matrix $\hat{T}_{r\rightarrow t}$, and the target view $I_t$ as input. In order to train the Indoor GeoNet in a weakly supervised manner, we define a total loss function $\mathcal{L}_{T}$ as the weighted summation of multiple losses as:
\begin{equation}
    \mathcal{L}_{T} = \sum_{(r,t)} \lambda_{P}\mathcal{L}_{P} + \lambda_{D}\mathcal{L}_{D} + \lambda_{C}\mathcal{L}_{C} + \lambda_{W}\mathcal{L}_{W}
\end{equation}
where $\mathcal{L}$'s are different loss functions explained in the following sections, $\lambda$'s are the corresponding loss weights, and $(r,t)$ iterates over all possible pairs of reference $I_r$ and target $I_t$ frames.

\subsubsection{Photometric Loss: $\mathcal{L}_{P}$}

The DepthNet and PoseNet networks can be trained by minimizing the photometric loss between the synthesized view (warped target view) $\hat{I}_{t\rightarrow r}$ and reference frame $I_r$: 
\begin{equation}
    \mathcal{L}_{P} = \sum_{(r,t)}\sum_{p_r} F_{diss}\big(I_r(p_r), \hat{I}_{t\rightarrow r}(p_r)\big)
\end{equation}
where  $\hat{I}_{t\rightarrow r}(p_r)=I_t(p_t)$, with warping between $p_t$ and $p_r$ obtained from \eqnref{geometry}, and $F_{diss}(.)$ is a dissimilarity measure between reference and synthesized frame.
To obtain $I_t(p_t)$ for estimating the value of $\hat{I}_{t\rightarrow r}(p_r)$, we used the differentiable bilinear sampling mechanism proposed in the spatial transformer networks \cite{jaderberg2015spatial} that linearly interpolates the values of the 4 neighboring pixels
of $p_t$ to approximate $\hat{I}_{t\rightarrow r}(p_r)$ such that:
\begin{equation}
\begin{aligned}
    \hat{I}_{t\rightarrow r}(p_r) = \sum_{\overset{i\in\{t,b\}}{j\in\{l,r\}}} w^{ij} I_t(p^{ij}_t)
\end{aligned}
\end{equation}
where $w^{ij}$ is linearly proportional to the spatial proximity between $p_t$ and $p^{ij}_t$, and $\sum_{i,j}w^{ij} = 1$. As far as $F_{diss}(.)$, we adopted the differentiable photometric dissimilarity measure proposed in \cite{godard2017unsupervised}, which has proven to be successful in measuring perceptual image similarity, and handling occlusions:
\begin{equation}
\begin{aligned}
    &F_{diss}\big(I_r, \hat{I}_{t\rightarrow r}\big)=\\&\alpha \frac{1-\text{SSIM}\big(I_r, \hat{I}_{t\rightarrow r}\big)}{2}
    +(1-\alpha)\big\|I_r-\hat{I}_{t\rightarrow r}\big\|_1
\end{aligned}
\end{equation}
where SSIM denotes the structural similarity index \cite{wang2004image} and $\alpha$ is taken to be $0.85$.

\subsubsection{Depth Smoothness Loss: $\mathcal{L}_{D}$}
The $\mathcal{L}_{P}$ loss function defined in the previous section is non-informative in homogeneous (monotone) regions of the scene where multiple depth maps and relative poses can result in the same warping. Therefore, as proposed in \cite{yin2018geonet}, we used a depth map smoothness loss term $\mathcal{L}_{D}$ weighted per-pixel by image gradients in order to obtain coherent depth maps while allowing depth discontinuities on the edges of the image:

\begin{equation}
    \mathcal{L}_{D}=\sum_{p_r}|\nabla D_{r}(p_r)| . \big(\exp(-|\nabla I_{r}(p_r)|\big)^T
\end{equation}
where $\nabla$ is the vector differential operator.

\subsubsection{Forward-Backward Consistency Loss: $ \mathcal{L}_{C}$}
We applied a forward-backward consistency check as in \cite{yin2018geonet} to enhance our predictions. Pixels for which the forward and backward flows (obtained from target to reference warping and vice versa) disagree significantly are considered as possible occluded regions. Therefore, such pixels are excluded from both the photometric loss and the forward-backward flow consistency check, and are defined as $\mathbf{p_r}$ (see \eqnref{consistency}) . Let us denote $f_{r\rightarrow t}(p_r) = p_r-p_t^{\{D_{p_r}, \mathbf{T}\}}$ as forward flow ($p_t$ is computed using \eqnref{geometry}), and conversely $f_{t\rightarrow r}(p_r) = p_r^{\{D_t,\mathbf{T}^{-1}\}}-p_t$ as backward flow, and $\Delta f_{t,r}(p_r) = f_{r\rightarrow t}(p_r) - f_{t\rightarrow r}(p_r)$. Then, the geometry consistency is imposed by adding the following loss term:
\begin{equation}
\label{eqn:consistency}
\begin{aligned}
    \mathcal{L}_{C} &= \sum_{p_r\in \mathbf{p_r}} \big\|\Delta f_{t,r}(p_r)\big\|_1\\
    \text{such that}&\\
    \mathbf{p_r} &= \big\{p_r: \|\Delta f_{t,r}(p_r)\|_2<\max(\alpha, \beta \|f_{r\rightarrow t}(p_r)\|_2)\big\}
\end{aligned}
\end{equation}
in which $(\alpha,\beta)$ are set to be $(3.0, 0.05)$. 
Please note that both the photometric loss $\mathcal{L}_{P}$ and geometric consistency loss $\mathcal{L}_{C}$ are enforced on pixel locations in 
$\mathbf{p_r}$, where there is little contradiction between forward and backward flow.

\subsubsection{Weak Supervision Loss: $\mathcal{L}_{W}$}
In order to enhance the overall performance of the network in prediction of depth and camera pose, we enforced the groundtruth depth maps, $D^{gt}$, by introducing another loss term on pixel locations for which we have the groundtruth depth values:
\begin{equation}
    \mathcal{L}_{W} = \sum_{i\in r,t}\big\|D_i-D_i^{gt}\big\|_2
\end{equation}
where $D_i$ and $D^{gt}$ are the predicted and groundtruth depth maps of training samples, respectively.

\depth
\section{Experimental Results and Evaluation}
Here, we report the Indoor GeoNet performance in depth and pose estimation as well as novel view reconstruction evaluated using publicly-available RGBD indoor and outdoor datasets. We also compared the performance of our proposed weakly supervised model with the unsupervised version of the model trained on indoor datasets as well as the original GeoNet trained solely on an outdoor dataset.  

\subsection{Indoor GeoNet Architecture and Implementation Details}
The Indoor GeoNet contains two sub-networks, the DepthNet, and the PoseNet, which construct the novel view synthesis of a rigid scene by leveraging geometric constraints, similar to the original GeoNet structure \cite{yin2018geonet}. The DepthNet consists of an encoder part and a decoder part. The encoder part has the structure of ResNet50 \cite{he2016deep}
and the decoder part uses deconvolution layers to enlarge predicted depth maps to their original resolution (as input) in a multi-scale scheme. Several skip connections are used between encoder and decoder parts in order to reuse high level or detailed information that was captured in the initial layers for reconstruction process in deconvolution layers. 
The PoseNet has the same architecture as in \cite{yin2018geonet}, which consists of 8 convolutional layers followed by a global average pooling layer that outputs the 6-DoF camera poses including rotation and translation. We used batch normalization \cite{ioffe2015batch} and ReLU activation functions \cite{nair2010rectified} for all of the convolutional layers except the prediction layers.

We considered the training sequence length to be $N=5$, and resized all the RGB frames to $144\times256$ pixels, and then trained the network with learning rate of $0.0002$, and batch size of $4$ for $20$ epochs in Tensorflow \cite{abadi2016tensorflow}. We set the loss weights to be $\lambda_P= \lambda_W =1, \lambda_D=0.5, \lambda_C=0.2$, and used Adam optimizer \cite{kingma2014adam} with its parameters set as $\beta_1 = 0.9, \beta_2 = 0.999$ for network training.

\subsection{Evaluation Datasets}
We performed the performance evaluation and comparison of our Indoor GeoNet using four available indoor and outdoor scene datasets: MobileRGBD, RSM Hallway, and KITTI raw and odometry datasets. Some RGB and depth samples from each datasets are shown in the first and second columns of \figref{depth}.

MobileRGBD is a corpus dedicated to low-level RGBD algorithms benchmarking on mobile platform. This dataset contains RGB and depth videos taken from a multi-section hallway scene with a moving robot equipped with a Microsoft Kinect v2. Each path is taped several times at different robot angles forming a total of 25 videos each with duration less than 1 minute. The robot information including the odometry data from robot (location coordinates internally reported by robot), and control commands to the robot (linear and angular velocity and stop commands) are also included within this dataset. This information makes this dataset suitable for our evaluation since it gives synchronized RGB, depth, pose and location information. We preprocessed MobileRGBD dataset in order to register images from RGB and depth cameras, since they have different dimensions, aspect ratio, fields of view, and cameras are positioned some distance apart from each other. The resolution for RGB image is $1920\times1080$, however, for depth image is $512\times412$. Therefore, we registered depth images to their RGB counterparts, and warped them into the same size. We preprocessed all of the images to sequences of $5$ consecutive frames with $144\times256$ resolution, forming a total of roughly $3200$ training samples. We left two videos for evaluation of network performance.

RSM Hallway dataset includes videos from hallways of RSM building at the Imperial College London, which can be a proper dataset for our indoor training purposes. This dataset contains videos from 6 hallways, each taken 10 times forming a total of 60 videos with $1280\times720$ RGB resolution. Similar to RGBD dataset, we preprocessed all of the videos to sequences of $5$ consecutive frames with $144\times256$ resolution, forming a total of roughly $18000$ training examples. However, we excluded videos from the first hallway for evaluation purposes.

The KITTI raw and odometry datasets are collected with two high-resolution color and gray-scale video cameras mounted on top of a standard station wagon while it is driving around a city, in rural areas and on highways. For the KITTI raw dataset, accurate groundtruth values for depth is provided by a Velodyne laser scanner. The KITTI odometry dataset consists of 22 stereo sequences, with half of the sequences having groundtruth trajectories and camera pose, which makes it suitable for outdoor camera pose estimation approaches.

\tablecomparedepth

\subsection{Estimation Performance Evaluation}

The weakly supervised Indoor GeoNet (referred to as IndoorGeoNet-WSup) performance is evaluated against two other models, one the original GeoNet trained on KITTI raw and odometry datasets (referred to as GeoNet-UnSup) and the other one the GeoNet trained from scratch in an unsupervised manner on the RSM Hallway dataset (referred to as IndoorGeoNet-UnSup). The performance comparison is done in two aspects: (1) the accuracy of depth and camera pose estimation using different approaches, (2) the reconstruction accuracy of the novel RGB scene synthesis using different approaches. The quantified results are calculated and reported for the datasets with available groundtruth depth and pose labels. Please note that we have both groundtruth depth and camera pose for MobileRGBD dataset, groundtruth depth for KITTI Raw dataset, groundtruth camera pose for KITTI Odometry dataset, while no depth or pose groundtruth for RSM Hallway is available. Based on the availability of the groundtruth depth labels, we chose  a subset of MobileRGBD datasets (9 out of 21 videos) in the weak supervision of IndoorGeoNet-WSup initialized by the IndoorGeoNet-UnSup network trained on RSM Hallway dataset.

\subsubsection{Depth and Pose Estimation}
We computed the depth and relative camera pose root mean squared error (RMSE) on the test set for those datasets/sequences for which we have the groundtruth values, and reported the errors in \tblref{depth} for the three models. Although GeoNet-UnSup works pretty well on the KITTI Raw and Odometry datasets, its performance degrades significantly on indoor datasets compared to the IndoorGeoNet-WSup, proving that the model is not generalizable to the indoor scenes. We also depicted some sample figures of depth prediction for the three models side by side in \figref{depth} along with groundtruth depth maps (if available) for comparison using sample monocular images from MobileRGBD, RSM Hallway, and KITTI datasets. As seen in this figure, although GeoNet-UnSup predicts satisfactory depth maps for the KITTI dataset, its predicted depth maps for MobileRGBD and RSM Hallway samples, hardly give any information about the general geometry of the scene, and the edges are completely lost. On the other hand, IndoorGeoNet-UnSup gives a fair prediction of the global geometry of the scene on sample images of RSM Hallway (on which the model is trained), however, predicted depth maps completely miss the details. The predictions of this model on the MobileRGBD sample images (not seen by the model during training) show that this model also fails to adapt to a new unseen indoor scene.

As seen in \tblref{depth}, with IndoorGeoNet-WSup model, depth and pose errors drop significantly for MobileRGBD dataset as compared to other models, since we are adding the depth supervision. Its predicted depth maps on sample images of MobileRGBD dataset clearly demonstrates the effect of supervision (even weak) on the ability of the network to learn detailed maps as shown in \figref{depth}. IndoorGeoNet-WSup also shows acceptable depth image estimation on other indoor scene (e.g. RSM Hallway) that were not part of weak supervision. This demonstrates the generalization capability of the proposed IndoorGeoNet-WSup.    

\subsubsection{Novel View Reconstruction Estimation}
Similar to the case of depth and pose estimation evaluation, we first check the adaptability of original GeoNet-UnSup to the indoor scenes of the RSM Hallway and MobileRGBD datasets. The mean image photometric loss $\mathcal{L}_{P}$, plus the structural similarity index measure between the reference image and the inverse warped target image are reported in table \tblref{reconst} for all of the dataset on our three models. Evident from this table, for GeoNet-UnSup, the reconstruction loss increases significantly on the MobileRGBD and RSM Hallway datasets that are not seen by the network during the training, which proves that the network fails to adapt to the indoor scenes. IndoorGeoNet-UnSup gives the lowest reconstruction $\mathcal{L}_{P}$ loss on RSM Hallway dataset (on which its network is trained), however, IndoorGeoNet-WSup also gives a comparable SSIM on this dataset, although it has not seen the dataset during the training. As expected, IndoorGeoNet-WSup gives the best reconstruction results on test set of MobileRGBD dataset with which the network is trained in a weak supervision fashion.
We also depicted some sample images of novel view reconstruction on MobileRGBD, RSM Hallway, and KITTI datasets using the three models in \figref{const}. As seen in this figure, IndoorGeoNet-WSup is able to successfully reconstruct novel nearby view from input images on both sample images of MobileRGBD and RSM Hallway where both depth and pose predictions contribute to the loss. Although GeoNet-UnSup works well on KITTI sample images, it fails to correctly reconstruct the novel view of indoor scenes. Using the IndoorGeoNet-UnSup model, the reconstructed views of RSM Hallway sample images are acceptable, because as we discussed in the previous section, its predicted depth maps give a fair estimation of the global geometry of the scene.

\reconst

\tablecomparerecons

\section{Conclusion}
\label{sec:conclusion}
In this work, we presented a weakly supervised hybrid learning approach to estimate depth data and relative camera pose targeted for indoor scenes. Our approach is inspired by the recent works in unsupervised scene understating, which integrate the  scene geometry constraints into the deep learning frameworks. However, the state-of-the-work in this domain are mostly concentrated on outdoor scenes found in datasets such as KITTI and CityScape. Here, we argued that these approaches are harldy transferable to indoor scenes, there is much more variations and more precision with detailed information is needed. In contrast, we proposed ``Indoor GeoNet'' using a weak supervision in terms of depth to improve both depth and pose predictions for indoor datasets. We believe that such supervision is sensible due to the availability of  inexpensive indoor RGB and depth sensors and several open-source indoor datasets. We compared the outcomes of our Indoor GeoNet in terms of depth, camera pose and novel view  estimation with the original unsupervied GeoNet models trained on different benchmark datasets. The results revealed that Indoor GeoNet is able to detect more detailed depth maps and also the pose estimation is improved when applied on indoor datasets.


{\small
\bibliographystyle{ieee}
\bibliography{paper}
}
\end{document}